\documentclass[sigconf]{acmart}

\usepackage{booktabs} % For formal tables
\usepackage{multirow}
\usepackage{subfigure}
\usepackage{graphicx}
\usepackage{url}
\usepackage[bf, medium]{titlesec}
\usepackage{balance}

\setcopyright{none}
\begin{document}
\title{Generating Video Descriptions with Topic Guidance}

\author{Shizhe Chen}
\affiliation{%
  \institution{Multimedia Computing Lab\\School of Information\\Renmin University of China}
}
\email{cszhe1@ruc.edu.cn}

\author{Jia Chen}
\affiliation{%
  \institution{Language Technologies Institute\\School of Computer Science\\Carnegie Mellon University}
}
\email{jiac@cs.cmu.edu}

\author{Qin Jin}
\authornote{Corresponding author.}
\affiliation{%
  \institution{Multimedia Computing Lab\\School of Information\\Renmin University of China}
}
\email{qjin@ruc.edu.cn}

\begin{abstract}
Generating video descriptions in natural language (a.k.a. video captioning) is a more challenging task than image captioning as the videos are intrinsically more complicated than images in two aspects. 
First, videos cover a broader range of topics, such as news, music, sports and so on. 
Second, multiple topics could coexist in the same video. 
In this paper, we propose a novel caption model, topic-guided model (TGM), to generate topic-oriented descriptions for videos in the wild via exploiting topic information.  
In addition to predefined topics, i.e., category tags crawled from the web, we also mine topics in a data-driven way based on training captions by an unsupervised topic mining model. 
We show that data-driven topics reflect a better topic schema than the predefined topics. 
As for testing video topic prediction, we treat the topic mining model as teacher to train the student, the topic prediction model, by utilizing the full multi-modalities in the video especially the speech modality. 
We propose a series of caption models to exploit topic guidance, including implicitly using the topics as input features to generate words related to the topic and explicitly modifying the weights in the decoder with topics to function as an ensemble of topic-aware language decoders. 
Our comprehensive experimental results on the current largest video caption dataset MSR-VTT prove the effectiveness of our topic-guided model, which significantly surpasses the winning performance in the 2016 MSR video to language challenge.
\end{abstract}


\maketitle

\section{Introduction}
% importance of the task and applications
It is an ultimate goal of video understanding to automatically generate natural language descriptions of video contents.
A wide range of applications can benefit from it such as assisting blind people, video editing, indexing, searching or sharing. 
Drawing on the recent success of image captioning \cite{DBLP:conf/icml/LebretPC15,DBLP:conf/cvpr/VinyalsTBE15,xu2015show,you2016image}, where a sentence is generated to describe the image content, more researchers are paying attention to the video captioning task to translate videos to natural language. 
\begin{figure} \centering 
\includegraphics[width=1\linewidth]{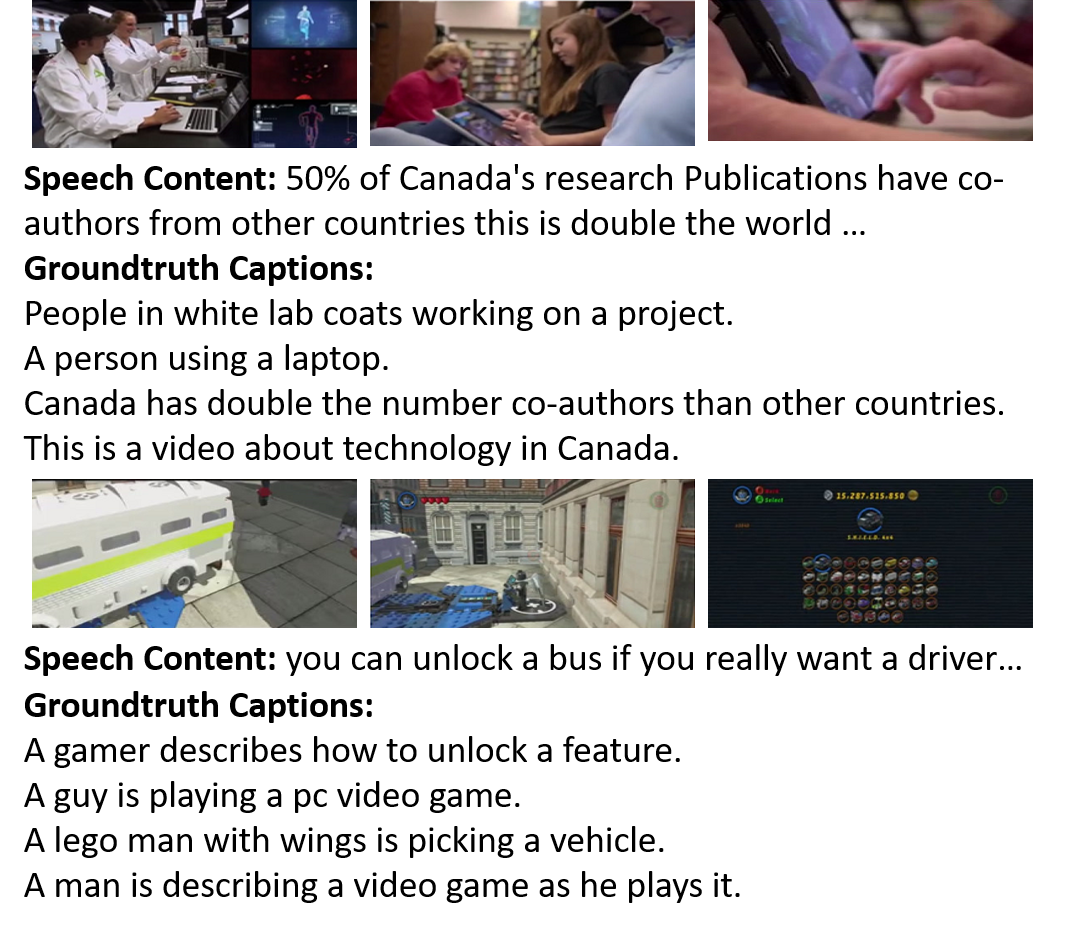} 
\caption{Topic diversity across videos and within one video.} 
\label{fig:caption_variety} 
\end{figure}

% the difficulties of video captioning: diverse contents (topics)
However, the open domain videos are quite diverse in topics which makes generating video descriptions more complicated than the image captioning.
For various topics ranging from political news to edited movie trailers, the vocabularies and expression styles vary a lot in describing the video contents.
For example, for political news videos, words from the political domain occur more frequently. 
Also political news descriptions are typically in the style of somebody reporting something, which are quite different from descriptions for other topics such as gaming, travel, movie, and animals etc.
Besides the topic diversity across videos, even in the same video, its diversity in content and video structure can result in very different video descriptions capturing different topics in the video as shown in Figure~\ref{fig:caption_variety}.
Therefore, the topic information is important to guide the caption model to generate better topic-oriented language expression. 

\begin{figure*} \centering 
\includegraphics[width=0.75\linewidth]{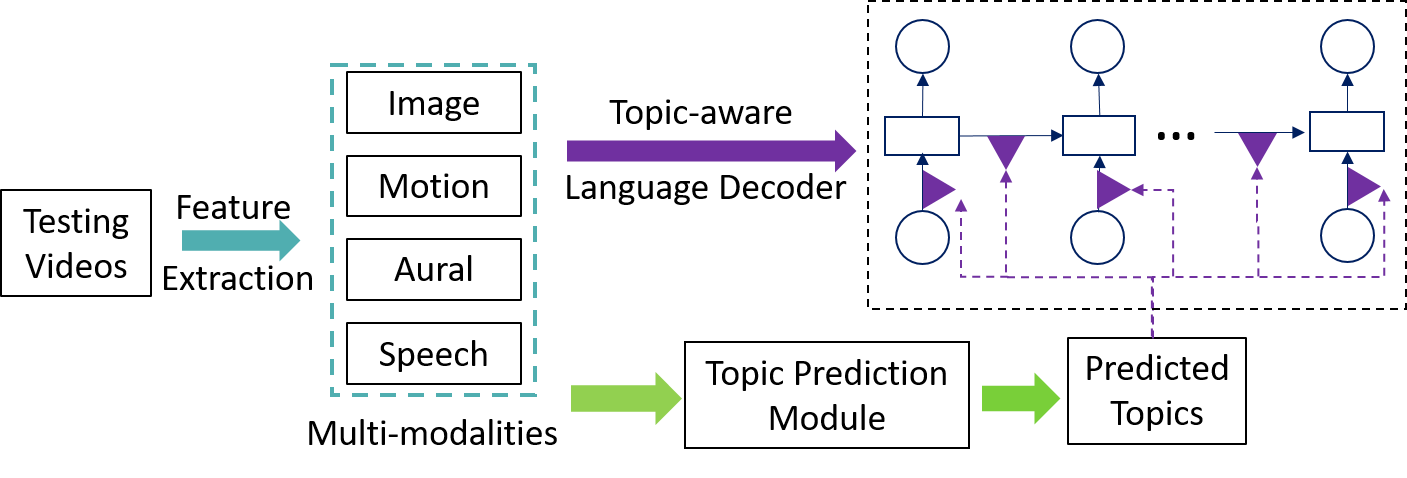} 
\caption{Framework of the proposed topic-guided model (TGM). We use the automatically mined topics as our topic guidance. For testing videos, multimodal features are utilized to predict the mined topics. Then the predicted topics are adopted in the topic-aware language decoder to generate topic-oriented video captions.}
\label{fig:framework} 
\end{figure*}

In our previous study \cite{DBLP:conf/mm/JinCCXH16}, we have utilized the predefined topics, the category tags crawled from video meta-data during data collection, to improve the captioning performance.
However, the predefined topics are suboptimal for video captioning because: 
1) The crawled information contains labelling mistakes which harms the captioning performance; 
2) The exclusive topic labels do not capture the topic diversity nature inside the video; 
3) The predefined topic schema is not specially designed for the video captioning task, which may not reflect the topic distributions well.
In this work, we propose the data-driven topics to deal with the drawbacks of predefined topics.
We use the Latent Dirichlet Allocation (LDA) topic mining model to automatically generate topics from the annotated video captions in training set.

In order to use the mined topic information in the captioning task, two questions need to be addressed:
1) how to obtain the topics automatically for testing videos;
and 2) how to effectively employ the topic information to model the difference of sentence descriptions in different video topics.

For the first question, we take a teacher-student learning perspective \cite{Ba2013Do} to train the data-driven topic prediction model.
The LDA topic mining model is viewed as the teacher to guide the student topic prediction model to learn.
A video generally contains multiple modalities including image, motion, aural and speech modalities.
Image modality provides rich information for understanding the video's semantic contents such as object and scene. The motion modality presents actions of the objects and the temporal structure of videos.
Aural and speech modalities provide additional information for understanding semantic topics from the sound perspective.
Hence, we build one general topic prediction model that utilizes all the four modalities and another topic prediction model dedicated to speech modality as its feature representation is different from other modalities. 

For the second question, we propose a novel topic-guided model (TGM) to employ the predicted topics, which is based on the encoder-decoder framework \cite{bahdanau2014neural}.
The TGM functions as an ensemble of topic-aware language decoders to learn specific vocabularies and expressions for various video topics.
We also compare the TGM with a series of caption models that we propose in this paper to exploit the topic information, including topic concatenation in encoder (TCE), topic concatenation in decoder (TCD), topic embedding addition/multiplication in decoder (TEAD/TEMD). 
These compared models implicitly use topics to change the input features of the encoder or decoder, while our proposed TGM explicitly modifies the weights in the decoder according to the predicted topics to capture the sentence distributions within the topic more effectively.
The framework of the overall system for testing videos is shown in Figure~\ref{fig:framework}.
Experimental results on the MSR-VTT dataset demonstrate the effectiveness of our proposed method, which can generate more comprehensive and accurate video descriptions.

In summary, our contributions in this work include: 
1) we show that the data-driven topics are more suitable as the topic representation for video captioning than the predefined topics, e.g. the category tags, with respect to the topic accuracy and schema; 
2) to the best of our knowledge, we are the first to use the full multi-modalities especially the speech modality to successfully boost the video captioning performance;
and 3) the proposed topic-guided model can exploit the topic information more effectively to generate better topic-oriented video descriptions.

The rest of the paper is organized as follows: Section 2 introduces the related work. Section 3 compares the predefined and the data-driven topics. Our proposed topic-guided model is described in Section 4. Section 5 presents experimental results and analysis. Section 6 draws some conclusions.
\section{Related Works}
There are mainly two directions in previous image/video captioning works.
The first is to build rule based systems, which first detect words by object or action recognition and then generate sentences with predefined language constrains.
For example, Lebret et al. \cite{DBLP:conf/icml/LebretPC15} predict phrases with a bilinear model and generate descriptions using simple syntax statistics.
Rohrbach et al. \cite{DBLP:conf/iccv/RohrbachQTTPS13} use the Conditional Random Field to learn object and activity labels from the video.
Such systems suffer from the expression accuracy and flexibility.

More recently, researches have been focusing on the second direction of encoder-decoder framework \cite{bahdanau2014neural} which generates sentences based on image/video features in an end-to-end manner.
For example, Vinyals et al. \cite{DBLP:conf/cvpr/VinyalsTBE15} utilize the LSTM to generate sentences with CNN features extracted from the image.
Venugopalan et al. \cite{Venugopalan2014Translating} transfer knowledge from image caption models with the encoder to perform mean pooling over frame CNN features for video captioning.
Pan et al. \cite{DBLP:conf/cvpr/PanMYLR16} explicitly embed the sentences and the videos into a common space in addition to the video description generation.

There are also works considering to employ semantic concepts in the encoder-decoder framework. 
For example, Wu et al. \cite{wu2016value} directly generate image captions based on the detected semantic concepts.
You et al. \cite{you2016image} propose to selectively attend to concept proposals in the decoder. 
Gan et al. \cite{Gan2016Semantic} propose the semantic compositional networks (SCN) which works as an ensemble of concept-dependent language decoder.
Our topic-guided model is inspired by SCN but with different aims of producing topic-oriented descriptions to address the topic diversity in video captioning. 
The reasons of using topics rather than semantic concepts are as follows:
1) There are much more objects in a video than in an image but many of them might be irrelevant to the video description. %So using the vague topics might be better than the semantic concepts;
2) The topics contain more additional information than semantic objects such as from the motion, aural and speech modalities;
and 3) The prediction accuracy is very important for the model as shown in \cite{Gan2016Semantic}. The topic classification is much easier than object classification.

For video captioning, various video topics result in quite diverse expressions compared with image captioning.
Previous works have explored generating descriptions for narrow-domain videos such as YouCook \cite{DBLP:conf/cvpr/DasXDC13} and TACoS \cite{DBLP:journals/tacl/RegneriRWTSP13}, whose vocabularies and expressions are similar in the dataset.
However, for open-domain videos with various topics such as the MSR-VTT dataset \cite{xu2016msr}, Jin et al. \cite{DBLP:conf/mm/JinCCXH16} exploit the predefined video categories in the encoder and significantly improve the captioning performance, which results in their winning of the MSR video to language challenge \cite{VTTGC2016}.
In our work, we further analyze the qualities of the predefined categories and propose to mine topics in a data-driven approach that leads to better accuracy and topic schema.

Multi-modality nature is also emphasized in video captioning.
For the motion modality in videos,
Yao et al. \cite{DBLP:conf/iccv/YaoTCBPLC15} explore the temporal structure with local C3D features and global temporal attention mechanism. 
Venugopalan et al. \cite{venugopalan2015sequence} propose the sequence to sequence structure which utilizes the LSTM as encoder to capture the temporal dynamics of videos.
Pan et al. \cite{pan2015hierarchical} further propose the hierarchical RNN  encoder as well as the temporal-spatial attention.
Aural modality has also been explored for video captioning. 
Jin et al. \cite{jin2016video,DBLP:conf/mm/JinCCXH16} and Ramanishka et al. \cite{ramanishka2016multimodal} integrate the visual and aural features in the encoder by early fusion and show that the multimodal fusion was beneficial to improve captioning performance.
In our work, we consider more modalities in videos especially for the use of speech content modality.

\begin{figure} \centering 
\includegraphics[width=1\linewidth]{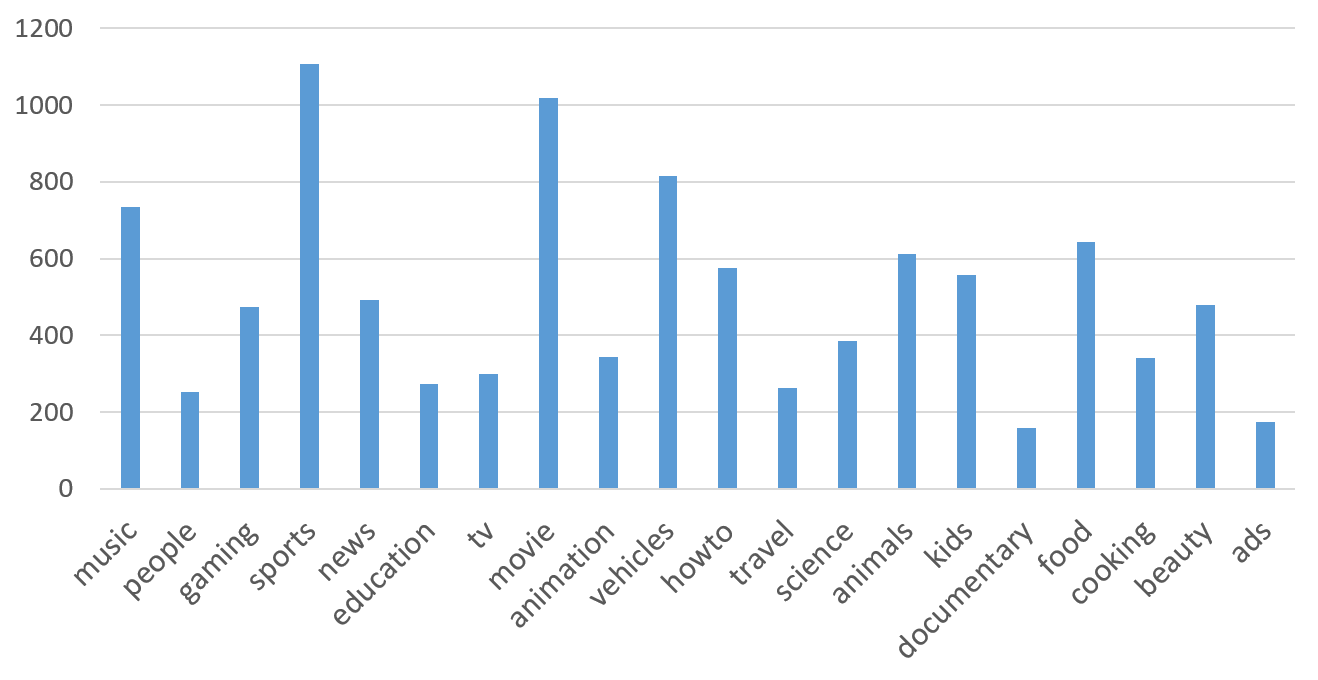} 
\caption{The number of video clips for the 20 predefined video categories of MSR-VTT dataset.} 
\label{fig:vtt_ctg} 
\end{figure}

\begin{figure} \centering 
\includegraphics[width=1\linewidth]{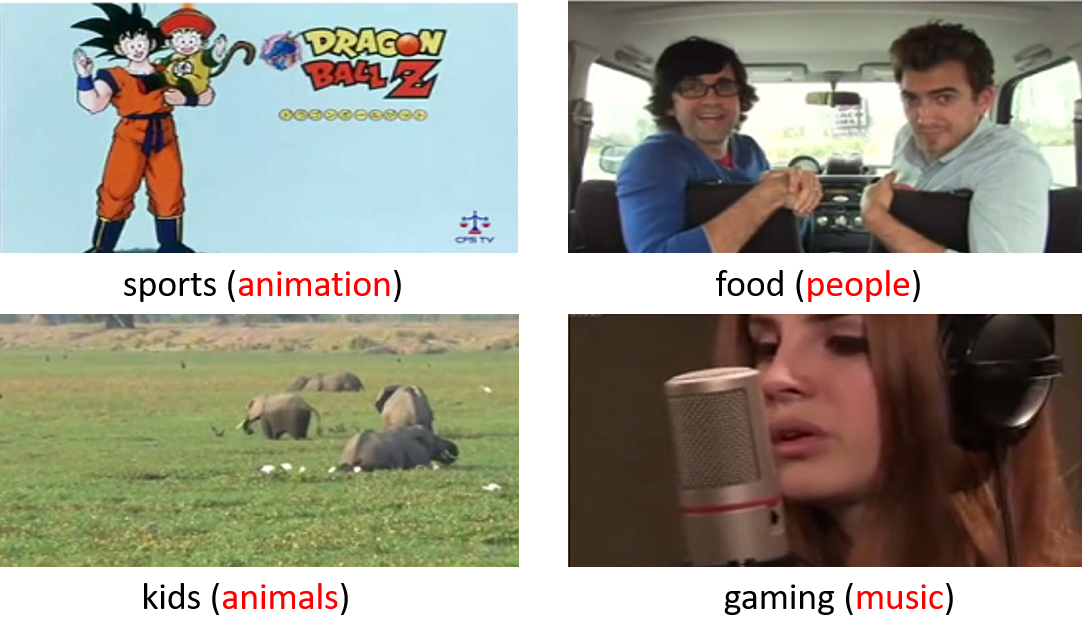} 
\caption{Examples of inaccurate category tags in MSR-VTT dataset. The word in black is the original category tag and the red is the more appropriate tag.} 
\label{fig:inaccurate_ctg} 
\end{figure}

\begin{table*}
\centering
\caption{Examples of some data-driven topics with their representative words and their co-occurrence with predefined category tags.}
\label{tab:topic_show}
\begin{tabular}{c|c|c|c}
topic id & \#videos & co-occurrence with predefined categories & representative words\\ \hline
1 & 182 & music:43\%, people:21\% & people dancing group girls dance women music video dances\\
3 & 281 & music:50\% & stage man playing singing band performing song guitar music\\
13 & 439 & food:63\%, cooking:29\% & food cooking kitchen dish person bowl ingredients pan preparing\\
8 & 864 & news:15\%, edu:14\%, sci:13\% & man talking talks guy speaking person sitting giving camera\\
\end{tabular}
\end{table*}

\section{Predefined vs. Data-driven Topics}
\label{sec:topic}
Our previous study \cite{DBLP:conf/mm/JinCCXH16} has shown that using predefined topics such as category tags can significantly boost video captioning performance.
In this section, we analyze the qualities of these predefined topics, and propose a data-driven approach to develop better topics for the captioning task.

\subsection{Predefined Topics: Category Tags}

Each video clip in the MSR-VTT dataset contains a predefined category tag derived from the meta-data of the video. 
The distribution of the category tags is shown in Figure~\ref{fig:vtt_ctg}.
The predefined category tags reflect the variety of the video topics, but there are mainly three disadvantages of them:

(1) Inaccurate category labels: The predefined category labels contain a certain amount of labelling mistakes as shown in the examples in Figure~\ref{fig:inaccurate_ctg}, which greatly harm the captioning performance.

(2) Exclusive topic distributions: The users can only assign one of the category labels. Such one-hot topic representation cannot reflect the topic diversity inside the video.

(3) Suboptimal topic schema: a) Ambiguous category definition. For example, the `people' category is too general to classify. b) Overlap between different categories. For example, the `food' and `cooking' categories cover almost similar videos. c) Large mixed categories. Some categories contain much more videos than others and are mixed with many subclasses. d) Indirect connection with captioning task. The category tags are defined to organize videos in the wild, they are not specifically defined for video captioning.

Therefore, although the predefined category tags have benefited video captioning a lot, there is still much room for improvement by defining a better topic schema to represent the diversity of videos for the  video captioning task.

\subsection{Data-driven Topics}
\label{subsec:minedtopics}
In order to overcome the drawbacks of the predefined category tags, we propose a data-driven way to generate a more suitable set of video topics. 
The human generated groundtruth captions provide us with rich and accurate annotations about the videos, which also reflect a more task-related topic distributions of the videos.
Thus, we propose to mine topics from the groundtruth video captions in the training set. 
We note that it requires no additional labelling effort on the dataset. 

% \textbf{Topics Mining:}
We observe that the multiple human generated groundtruth captions sometimes do not agree with each other even for the same video.
% jiac: csz add example, may be illustrated by a figure
For example, Figure~\ref{fig:caption_variety} shows an example video and its groundtruth captions which describe the video from different aspects including detailed frame contents, speech contents and general video contents.
Such example indicates that a video usually contains several topics, which are reflected in its multiple diverse groundtruth captions. 
The above observation aligns with the generation process of Latent Dirichlet Allocation (LDA) model \cite{DBLP:journals/jmlr/BleiNJ03}:\\*
1. the model first draws a topic index $z_{di} \sim Multinomial(\theta_{d})$ from the video, where $\theta_{d} \sim Dirichlet(\alpha), d=1,...,D$. \\*
2. the model draws the observed word $w_{ij} \sim Multinomial(\beta_{z_{di}})$ from the selected topic, where $\beta_{k} \sim Dirichlet(\eta), k=1...K$. \\*
Here, we group the multiple groundtruth captions of a video into one document to mine latent topics from the training data.
Stopwords are removed and the the bag-of-words representation is used as our document feature. 
% We train the LDA with the number of topics $K$ to be 20 on the training set in order to compare with the predefined 20 category tags.
% for each word $i$ in document $d$ (video's multiple groundtruth captions), 

% We exploit the Latent Dirichlet Allocation (LDA) model \cite{DBLP:journals/jmlr/BleiNJ03} to generate latent topics.
% The LDA model assumes a generative process for a corpus with $D$ documents and $K$ topics.
% For each word $i$ in document $d$, the model first draws a topic index $z_{di} \sim Multinomial(\theta_{d})$ and then draws the observed word $w_{ij} \sim Multinomial(\beta_{z_{di}})$, where $\theta_{d} \sim Dirichlet(\alpha), d=1,...,D$ and $\beta_{k} \sim Dirichlet(\eta), k=1...K$.
% Its graphical model is shown in the topic mining module of Figure~\ref{fig:topic_prediction}.
% After optimizing the joint probability distributions of the nodes, we can infer the latent topics of the documents. 

% \textbf{Qualities of the mined topics:}
\subsection{Relation between Predefined Topics and Data-driven Topics}
% jiac: what is compositional predefined category tags? 
% need to describe details of how to get this table
We study the relation between the predefined topics and data-driven topics based on their co-occurrence in videos. 
For each video in the training set, we have its corresponding predefined category tag and the data-driven topic distribution calculated from the LDA model. 
To simplify the calculation of co-occurrence, we assign each video with the most likely topic. 
Table~\ref{tab:topic_show} shows co-occurrence between some predefined categories and data-driven topics.
We can see that some predefined categories are split into different topics.
For example, the music category is mainly separated into topic 1 of dancing and topic 3 of singing.
Some content similar categories are combined together as one topic, i.e. topic 13 consisting of food and cooking categories.
And categories that are different but express similar content are also merged. 
For example, news, educations and science categories are merged into one as most of the descriptions under these categories are ``somebody is talking about something".
In summary, it shows that the data-driven topics are quite promising and reflect video content distributions better than predefined categories. 
% Therefore, qualitatively we consider the automatically mined topics are superior to the predefined category tags since they are more accurate and have a better topic schema related to the captioning task.
\section{Topic Guidance Model}
In this section, we provide our solutions for the following two problems: 
1) how to automatically predict topics for testing videos with multi-modalities;
2) how to maximize the effects of the topic information for caption generation.

\subsection{Multimodal Features}
\label{subsec:features}

We extract features from image, motion, aural and speech modalities to fully represent the content of videos.

\textbf{\emph{Image modality: }}
The image modality reflects the static content of videos. 
We extract activations from the penultimate layers of the inception-resnet \cite{Szegedy2016Inception} pre-trained on the ImageNet as image object features, and the penultimate layers of the resnet \cite{DBLP:conf/cvpr/HeZRS16} pre-trained on the places365 \cite{zhou2016places} as image scene features, the dimensionality of which are 1536 and 2048 respectively.

\textbf{\emph{Motion modality: }}
The motion modality captures the local temporal motion. 
We extract features from the C3D model \cite{tran2015learning} pre-trained on the Sports-1M dataset.
%The videos are segmented into non-overlap shots of 16 frames length. 
We extract activations from the last 3D convolution layer and max-pooling them along the spatial dimension (width and height) to obtain video features with dimensionality of 512. 
We then applied $l_{2}$-norm on the C3D features.

\textbf{\emph{Aural modality: }}
Aural modality is complementary to visual modalities, especially for distinguishing scenes or events. 
We extract the Mel-Frequency Cepstral Coefficients (MFCCs) \cite{davis1980comparison} as the basic low-level descriptors. 
Two encoding strategies, Bag-of-Audio-Words \cite{pancoast2014softening} and Fisher Vector \cite{sanchez2013image}, are used to aggregate MFCC frames into one video-level feature vector, with dimensionality of 1024 and 624 respectively. 
%The BoAW encoding is trained by K-means with 1024 centers and L1-norm is applied on the BoAW representations.
%The FV encoding contains 8 mixtures of the Gaussian Mixture Models (GMMs) and is normalized with L2-norm.

\textbf{\emph{Speech modality: }}
Speech modality provides semantic topics and content details of the video. 
We use the IBM Watson API \cite{ibm-watson-asr-api} for speech recognition. 
Since the backgrounds of the videos are noisy, we clean the speech transcriptions by removing transcriptions with less than 10 words and out-of-vocabulary words based on the training caption vocabulary.
Only about half portion of the videos contain speech transcriptions afterwards and a certain amount of transcription errors still exist in the transcriptions.
We use the bag-of-words representation as the speech modality feature.

\begin{figure} \centering 
\includegraphics[width=1\linewidth]{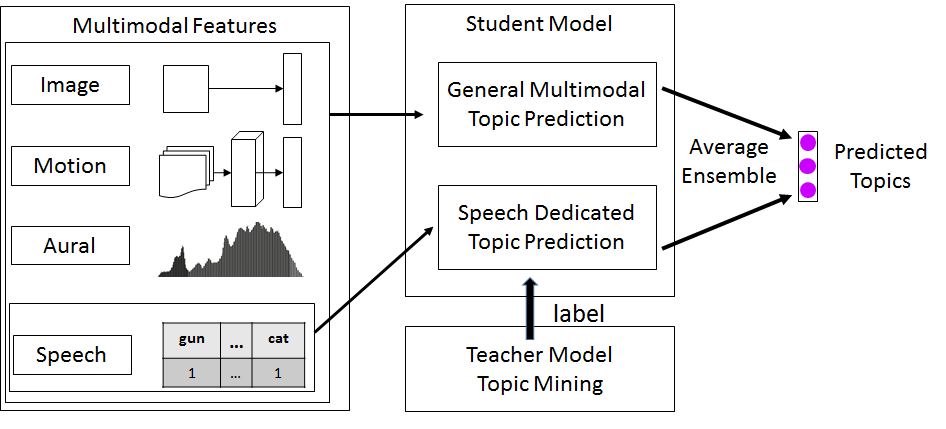} 
\caption{The framework for topic prediction. We treat the problem from a teacher-student perspective. The teacher topic mining model is used to guide the two student topic prediction models to learn based on general multi-modalilties and speech modality respectively.} 
\label{fig:topic_prediction} 
\end{figure}

\subsection{Topic Prediction}
\label{sec:topic_prediction}
% need better arrangement to introduce two types of transfers
For predefined topics, i.e. category tag, we train a standard one hidden layer neural network with cross-entropy loss to predict.
The inputs of this neural network are the multimodal features as described in the section~\ref{subsec:features}. 

For data-driven topics, as there is no direct topic class label, we leverage the topic distribution generated from the topic mining LDA model in section~\ref{subsec:minedtopics}. 
% jiac: csz to supplement reference for teacher-student learning
We take a teacher-student learning perspective \cite{Ba2013Do} to train the data-driven topic prediction model. 
To be specific, the topic mining LDA model is viewed as the teacher and the topic prediction model is viewed as the student. 
The teacher, topic mining model, is trained in unsupervised style and it generates label, i.e. topic, to guide the student, topic prediction model, to learn. 
First, we design two topic prediction models: one is general for all multimodal features and the other one is dedicated to speech modality features.
Then, we ensemble predictions from these two models by averaging to get the final prediction. 

% The mined topics are generated from the annotated captions, which are not available for testing videos.
% So we need to generate topic proposals based on the video representations.
% The topic mining module is viewed as the teacher for the topic prediction module from a student-teacher perspective.
% We propose two ways for knowledge transferring for the student learning procedure as shown in Figure~\ref{fig:topic_prediction}.
% For target transfer, we use the mined topics as the learning targets for our prediction networks,
% while for model transfer, the topic mining model is shared as the prediction model.

\textbf{General Multimodal Topic Prediction Model: }
% add hinton's dark knowledge paper
This model is designed to predict topic from the video content using all the multimodal features. 
In the teacher-student perspective, the student model usually learns the output distribution on labels, i.e. dark knowledge \cite{Hinton2015Distilling}, rather than output label from the teacher model. 
Following this way, we choose to use KL-divergence as our loss function. 
The formulation of KL-divergence is as follows:
\begin{equation}
D_{KL}(P||Q) = \sum_{k=1}^{K} P_{k} log \frac{P_{k}}{Q_{k}}
\end{equation}
where $P, Q$ are the probability distributions of the mined topics and predicted topics respectively. 
The $D_{KL}(P||Q)$ is differentiable and thus can be optimized via back-propagation.
% Our aim is to transfer the topic probabilities from the topic mining module to train the student networks to predict the mined topics.
% The feed forward neural networks are utilized as our prediction model.
% Although the commonly used classification loss is cross entropy, we choose to use KL-divergence as our loss function. 
% Because the KL-divergence imposes higher requirements that not only should the topics be correctly classified but also the probability distribution should be similar.
% The formulation of KL-divergence is as follows:
% \begin{equation}
% D_{KL}(P||Q) = \sum_{k=1}^{K} P_{k} log \frac{P_{k}}{Q_{k}}
% \end{equation}
% where $P, Q$ are the probability distributions of the mined topics and predicted topics respectively. 
% The $D_{KL}(P||Q)$ is differentiable and thus can be optimized via back-propagation.

\textbf{Speech Dedicated Topic Prediction Model: }
% jiac: note that this is only valid when text data is available
As speech modality is very informative but not always available in videos, we build a dedicate topic prediction model using speech features. 
% The topic mining model can infer the topic distribution for the text features,
% so it can be directly shared for speech texts to predict the mined topics.
Different from other modality features, the speech text feature is of high dimensionality with noises and is sparse. 
Instead of using the same architecture of the general multimodal topic prediction model, we design a very different architecture for the speech dedicated topic prediction model. 
We make a simple choice of the architecture: reusing the topic mining model for topic prediction on speech text features. 
As the representation of speech text features is the same as that of features used in the topic mining model, we don't need to reinvent the wheel. 
% Since the speech text features are of high dimensionality with noises and the data is sparse, the target prediction approach might easily result in over-fitting than the model sharing approach.
% Also, the probability distribution from the model sharing approach resembles the target ones better.

% \textbf{Integral Student Topic Prediction:}
% To integrate the different modalities and knowledge transferring strategies, we explore two kinds of fusion methods -- feature-level (early) fusion and decision-level (late) fusion \cite{wu2014survey}.
% The early fusion uses the concatenated features from different modalities as input features for classifiers,
% while the late fusion combines the predictions from different modalities and trains a second level model.
% In this work, the average of predictions is adopted as the second level model in the late fusion.

As to the problem of missing modalities, we only use videos that contain the corresponding speech modality in the speech dedicated topic prediction model. In general multimodal topic prediction model, the features of missing modalities are padded as zeros.
In ensembling, predictions of the missing modalities are not considered in the average process.

\subsection{Caption Models with Topic Guidance}
Suppose we have multiple video-sentence pairs $(V,\mathrm{y})$ in video captioning dataset, where $\mathrm{y}=\{w_1,\dots,w_{N_w}\}$ is the sentence with $N_w$ words.
Assume the multimodal features of the video are $m_{1}, \dots, m_{N_{m}}$, where $N_m$ is the number of modalities,
the multimodal encoder is a neural network that fuses the multimodal features into a dense video representation $\mathrm{x}$ as follows:
\begin{equation}
\mathrm{x} = W_{e}[m_1;\dots;m_{N_m}]+b_{e}
\end{equation}
where $W_{e},b_{e}$ are the parameters in the encoder and $[\cdot]$ denotes the feature concatenation.
Since the captioning output is the sequential words, we utilize the LSTM \cite{hochreiter1997long} recurrent neural networks as our language decoder:
\begin{equation}
h_t = f(h_{t-1}, w_{t-1}; \theta_d)\ \mathrm{for}\ t=1,\dots,N_w
\end{equation}
where $f$ is the LSTM update function, $h_t$ is the state of LSTM and $\theta_d$ is the parameter in LSTM.
We initialize $h_0$ as $\mathrm{x}$ to condition on the video representation and $w_0$ as the sentence start symbol.
Then the probability of the correct word conditioned on the video content and previous words can be expressed as:
\begin{equation}
\mathrm{Pr}(w_t|\mathrm{x},w_0,\dots,w_{t-1}) = \mathrm{Softmax}(W_d h_t + b_d)
\end{equation}
where $W_d,b_d$ are parameters. The objective function is to maximize the log likelihood of the correct description:
\begin{equation}
\log \mathrm{Pr(y|x)} = \sum_{t=1}^{N_w} \log \mathrm{Pr}(w_t|\mathrm{x},w_0,\dots,w_{t-1})
\end{equation}

We denote the predicted topics in Section~\ref{sec:topic_prediction} as $\mathrm{z} \in \mathbb{R}^{K}$, where $K$ is the number of topics.
In order to effectively exploit the topics $\mathrm{z}$ for video captioning, 
we propose the novel topic-guided model (TGM),
and a series of simple yet strong caption models with topic guidance for comparison, called TCE, TCD and TEAD/TEMD for short.
Detailed descriptions of these models are as follows.

\textbf{Topic Concatenation in Encoder (TCE): }
We fuse the topic distribution $\mathrm{z}$ together with multimodal features in the encoder as the topic-aware video representation $\mathrm{x}$:
\begin{equation}
\mathrm{x} = W_e([m_1;\dots;m_{N_m};\mathrm{z}]) + b_e 
\end{equation}
The LSTM decoder then generates video description conditioning on the new topic-aware representation $\mathrm{x}$.

% jiac: it is too difficult for me to figure out a place to place the following paragraph. So I just remove it. 
% As a preliminary experiment, we use the predefined categories with the TCE model to show the effects of the topic features.
% Figure~\ref{fig:ctg_effects} shows some generated video descriptions with or without the category feature.
% We can see that the category behaves like a guidance to adjust the generated sentence more related to the category. 
% So we consider the effect of the topic information is to assist the caption model with language biases to the topic, and thus using the topic information in the language decoder might provide stronger guidance to learn topic specific expressions.

%For example, the system without category guidance failed to recognize any animals in the video as shown in the bottom example in Figure~\ref{fig:ctg_effects}.
%But when equipped with the category information, the generated sentence is much more semantically relevant to the animal category though it is not perfect to recognize the exact animal.
%
%From the above analysis, we think that the provided topic information might assist the model with language biases to the topic of the video such as being able to produce related words belonging to the certain topic and thus improve the quality of the sentence generation.
%So explicitly using the topic information in the language decoder might provide stronger guidance to learn the different vocabularies and expression styles for different video topics.
%The following models are proposed to use topic features in the language decoder.

\textbf{Topic Concatenation in Decoder (TCD): }
In the TCE model, the topic information only occurs at the first step in the decoder, which could easily make the topic guidance ``drift away".
To enhance the topic guidance, we concatenate the topic distribution $\mathrm{z}$ with the word embedding $w_{t-1}$ as the input to the LSTM every step, which is similar to the gLSTM proposed in \cite{jia2015guiding}:
\begin{equation}
h_t = f(h_{t-1},[w_{t-1};\mathrm{z}];\theta_d)
\end{equation}
The extra input of topic $\mathrm{z}$ can guide the language decoder to generate words related to the topic in every step.

\textbf{Topic Embedding Addition/Multiplication in Decoder (TEAD/TEMD):}
To generate a more comparable representation for the topic representation $\mathrm{z}$ compared to the word embedding, we embed each topic into a latent vector space with the same dimensionality as the word embedding:
\begin{equation}
\mathrm{z}_e = W_z \mathrm{z} + b_z
\end{equation}
where $W_z,b_z$ are topic embedding parameters.
We perform addition or multiplication on the topic embedding and word embedding to generate the topic-aware input feature for the language decoder every step, which are expressed as:
\begin{align}
\mathrm{TEAD}:\ h_t &= f(h_{t-1}, w_{t-1} \oplus \mathrm{z}_e; \theta_d) \\
\mathrm{TEMD}:\ h_t &= f(h_{t-1}, w_{t-1} \odot \mathrm{z}_e; \theta_d)
\end{align}
where $\oplus, \odot$ are element-wise addition and multiplication respectively.

\textbf{Topic-Guided Model (TGM):}
The above TCD and TEAD/TEMD models only implicitly using the topic information as the global guidance, which modify the inputs to the language decoder in every step and cannot take into account of the overall expressions within the topic.
Inspired by Gan et al. \cite{Gan2016Semantic}, therefore, we further propose the topic-guided model (TGM) that explicitly functions as an ensemble of topic-aware language decoders to capture different sentence distributions for each topic. 
The structure of the TGM is shown in the right side in Figure~\ref{fig:framework}, which can automatically modify the weight matrices in LSTM according to the topic distribution $\mathrm{z}$.

Let us take one of weight matrices in the LSTM as an example, and other parameters in LSTM cell are alike.
We define weight $W_{\tau} \in \mathbb{R}^{n_{h}\times n_{w} \times K}$, where $n_{h}$ is the number of hidden units and $n_{w}$ is the dimension of word embedding.
The $W_{\tau}$ can be viewed as the ensemble of $K$ topic-specific LSTM weight matrices.
The topic-related weight matrix $W(\mathrm{z}) \in \mathbb{R}^{n_{h}\times n_{w}}$ can be specified as 
\begin{equation}
W(\mathrm{z}) = \sum_{k=1}^{K} \mathrm{z}_{k}W_{\tau}[k]
\end{equation}
where $\mathrm{z}_{k}$ is the $k$-th topic in $\mathrm{z}$; $W_{\tau}[k]$ denote the $k$-th matrix of $W_{\tau}$.
In this way, the video topic $\mathrm{z}$ can automatically generate its corresponding LSTM decoders to produce the topic-oriented video descriptions.
However, the parameters are increasing with $K$ which may result in over-fitting easily. 
So the ideas in \cite{Memisevic2007Unsupervised} are used to share parameters by factorizing $W(\mathrm{z})$ as follows:
\begin{equation}
W(\mathrm{z}) = W_{a} \cdot diag(W_{b}\mathrm{z}) \cdot W_{c}
\end{equation}
where $W_a \in \mathbb{R}^{n_{h}\times n_{f}}$, $W_b \in \mathbb{R}^{n_{f}\times K}$ and $W_c \in \mathbb{R}^{n_{f}\times n_{w}}$. $n_{f}$ is the number of factors.
$W_{a}$ and $W_{c}$ are shared among all topics, while $W_{b}$ can be viewed as the latent topic embedding.

%\begin{figure} \centering 
%\includegraphics[width=0.9\linewidth]{figs/category-effects} 
%\caption{Examples of generated sentences with and without category tags.} 
%\label{fig:ctg_effects} 
%\end{figure}

\section{Experiments}

\subsection{Experimental Setup}

\textbf{Dataset:}
The MSR-VTT corpus \cite{xu2016msr} is currently the largest video to language dataset with a wide variety of video contents. It consists of 10,000 video clips with 20 human generated captions per clip. Each video also contains a predefined category tag, which is one of the 20 popular video categories in web videos. Following the standard data split, we use 6,513 videos for training, 497 videos for validation and the remained 2,990 for testing.

\textbf{Data Preprocessing:}
We convert all descriptions to lower case and remove all the punctuations. 
We add begin-of-sentence tag $<$BOS$>$ and end-of-sentence tag $<$EOS$>$ to our vocabulary. 
Words which appear more than twice are selected, resulting in a vocabulary of size 10,868. 
The maximum length of a generated caption is set to be 30. 

\textbf{Training Settings:}
We empirically set the feed forward neural networks for topic prediction to have one hidden layer with 512 units.
The dimension of LSTM hidden size is set to be 512.
The output weights to predict the words are the transpose of the input word embedding matrix.
We apply dropout with rate of 0.5 on the input and output of LSTM and use ADAM algorithm \cite{kingma2014adam} with learning rate of $10^{-4}$. 
Beam search with beam width of 5 is used to generate sentences during testing process.
The baseline system is the vanilla encoder-decoder framework with multimodal features (we call it the multimodal baseline)\footnote{In our experiments, we find that the dimensionality of the features to the encoder should not be too high in order to avoid over-fitting, so only the image object, video motion, and aural features are used as input to the encoder.}.

\textbf{Evaluation Metrics:}
We evaluate the caption results comprehensively on all major metrics, including BLEU \cite{papineni2002bleu}, METEOR \cite{denkowski2014meteor}, ROUGE-L \cite{lin2004rouge} and CIDEr \cite{vedantam2015cider}.

\subsection{Evaluation}
Table~\ref{tab:ctg_topic_comparison} presents captioning performances on testing set with predefined category tags and the data-driven topics using their corresponding best caption model with topic guidance.
We can see that the different topic guidances (the second to forth rows) all greatly improve the performance of the multimodal baseline (the first row).
Since the data-driven topics on testing set are predicted as shown in Figure~\ref{fig:topic_prediction}, we also use the predicted category tags for a fair comparison.
The guidance from predicted data-driven topics outperforms that from the predicted category tags on all four evaluation metrics, and the Student's t-test shows the improvement is significant with p-value$<0.002$.
Even compared with the category tags assigned by video uploaders,
the predicted data-driven topics also slightly boost the captioning performance on multiple metrics with the Student's t-test p-value$<0.01$ on BLEU$@$4 and CIDEr metrics, which shows the performance gain is robust.
These results suggest that the data-driven topics are more suitable as the topic representation than predefined topics for video captioning.

%The category tag prediction model achieves prediction accuracy of 57.16\% on the testing set,
%while the data-driven prediction model achieves accuracy of 63.65\%.
%The higher prediction performance of the data-driven topics suggests that the data-driven topics are more coherent in video contents than the predefined category.
%The caption performance of predicted categories are inferior to that of groundtruth categories, which suggests the topic prediction performance is important for the topic guidance.
%However, the proposed TGM model with predicted topics improve the performance by a large margin compared to the predefined category tags on most of the metrics, which shows that the data-driven topics are superior to the predefined topics even with imperfect prediction.

\begin{table}
\centering
\caption{Captioning performance of the predefined categories and data-driven topics using the best caption model with topic guidance. The acronyms B, M, R, C denote BLEU$@$4, METEOR, ROUGE-L and CIDEr respectively.}
\label{tab:ctg_topic_comparison}
\begin{tabular}{c|c|c|c|c}
 & B & M & R & C\\ \hline
multimodal & 0.4211 & 0.2872 & 0.6170 & 0.4608 \\ \hline
pred category TGM & 0.4276 & 0.2885 & 0.6167 & 0.4754 \\ 
pred topic TGM & \textbf{0.4397} & \textbf{0.2921} & 0.6234 & \textbf{0.4970}\\ \hline
category TCE & 0.4343 & 0.2921 & \textbf{0.6249} & 0.4890\\ 
\end{tabular}
\end{table}

To demonstrate the effectiveness of the proposed topic-guided model (TGM), we further compare the TGM with other caption models with topic guidance.
As shown in Table~\ref{tab:topic_enhanced_models}, the TGM achieves the best performance among all the caption models on all four metrics especially for the CIDEr score.
It suggests that modifying the weights of the decoder according to the topic distributions can employ the topic guidance more effectively to generate better topic-oriented descriptions.
Our proposed TGM model also achieves better performance than the winning performance in 2016 MSR video to language challenge \cite{DBLP:conf/mm/JinCCXH16}, where we use multimodal features and select best models by the predefined categories. 

\begin{table}
\centering
\caption{Captioning performance comparison among caption models with topic guidance (using the predicted data-driven topics). The acronyms B, M, R, C denote BLEU$@$4, METEOR, ROUGE-L and CIDEr respectively.}
\label{tab:topic_enhanced_models}
\begin{tabular}{c|c|c|c|c}
 & B & M & R & C\\ \hline
pred topic TCE & 0.4353 & 0.2916 & 0.6226 & 0.4783\\ 
pred topic TCD & 0.4246 & 0.2861 & 0.6171 & 0.4714\\
pred topic TEAD & 0.4333 & 0.2893 & 0.6221 & 0.4765\\
pred topic TEMD & 0.4304 & 0.2884 & 0.6207 & 0.4694\\ \hline
pred topic TGM & \textbf{0.4397} & \textbf{0.2921} & \textbf{0.6234} & \textbf{0.4970}\\ \hline
v2t\_navigator \cite{DBLP:conf/mm/JinCCXH16} & 0.4080 & 0.2820 & 0.6090 & 0.4480 \\
\end{tabular}
\end{table}

\begin{figure} \centering 
\includegraphics[width=1\linewidth]{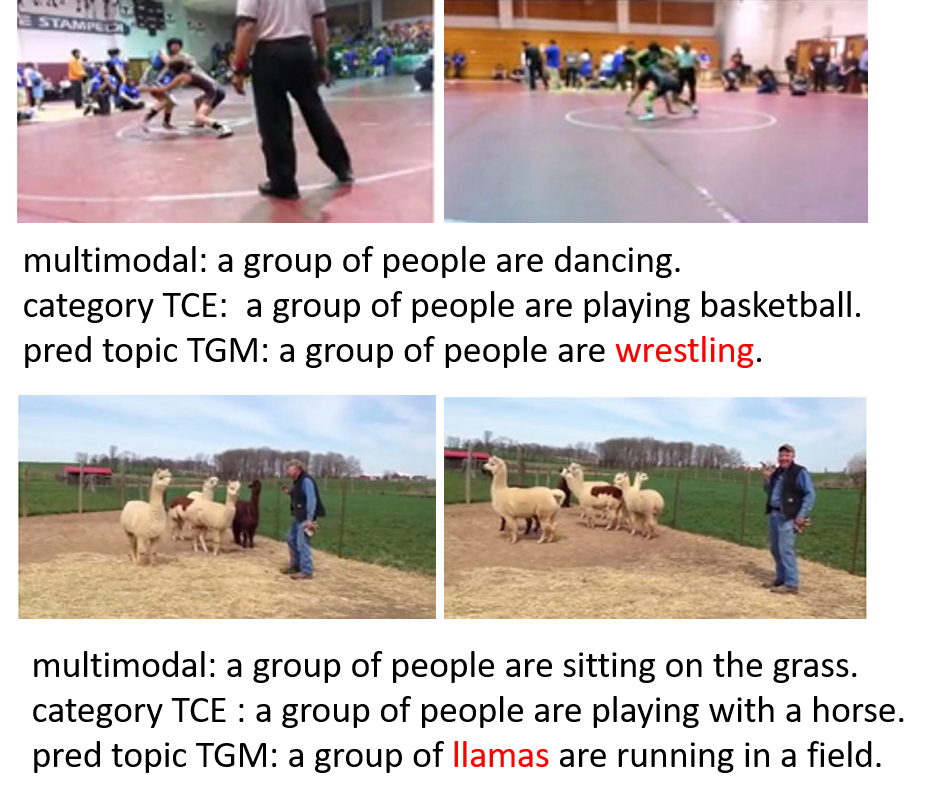} 
\caption{Examples on testing set to demonstrate the effectiveness of the TGM.} 
\label{fig:case_study} 
\end{figure}

Figure~\ref{fig:case_study} presents some examples in the testing set.
In addition to more accurate video descriptions, the TGM can also generate more novel concepts such as the llama.
Our statistics on the generated sentence show that the number of unique caption words generated by TGM is 397, while it is 360 by the multimodal baseline model.

\begin{table}
\centering
\caption{Captioning performance of predicted topics and the annotated topics. The acronyms B, M, R, C denote BLEU$@$4, METEOR, ROUGE-L and CIDEr respectively.}
\label{tab:annotate_topic_metric}
\begin{tabular}{c|c|c|c|c}
& B & M & R & C\\ \hline
pred topic TGM & 0.4397 & 0.2921 & 0.6234 & 0.4970\\ 
anno topic TGM & \textbf{0.4548} & \textbf{0.3009} & \textbf{0.6339} & \textbf{0.5418}\\
\end{tabular}
\end{table}

\begin{figure} \centering 
\includegraphics[width=1\linewidth]{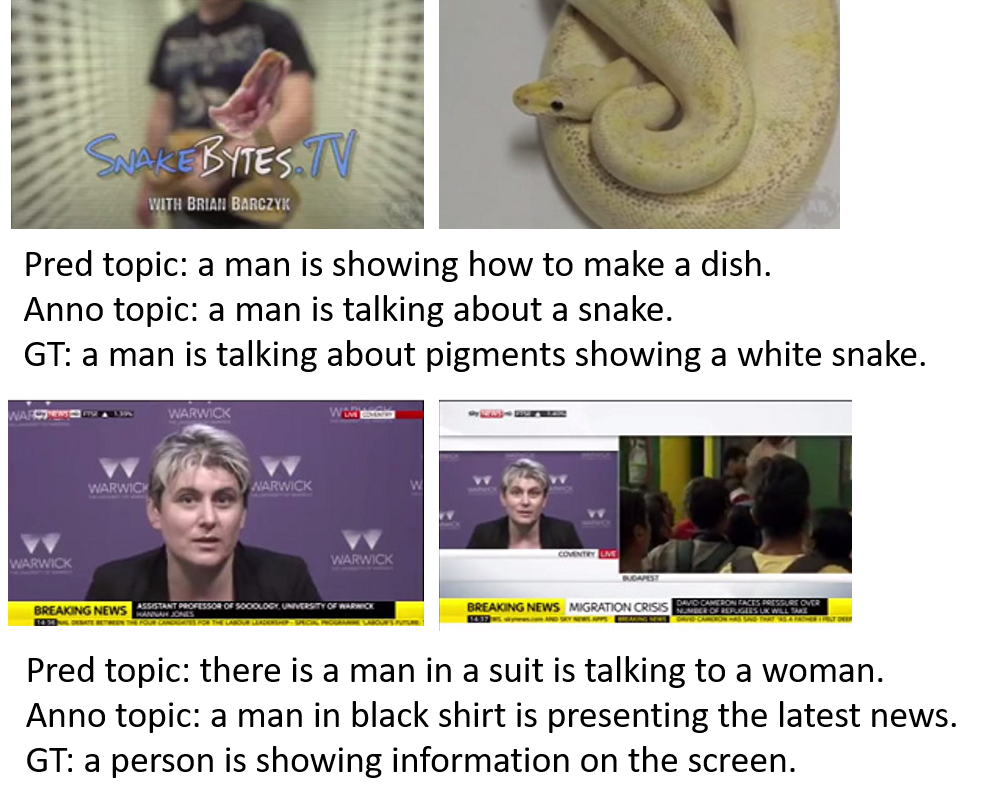} 
\caption{Examples on testing set. Pred topic and anno topic denote using the predicted and annotated topics in the topic-guided model. GT is a random groundtruth caption sentence.} 
\label{fig:gt_topic_eg} 
\end{figure}

\subsection{Ablation Experiments}
\textbf{Interactive Caption with Manually Annotated Topic:}
Our model offers the flexibility of manually assigned topics to the video. 
It means that we could interactively annotate the topics for testing videos based on relevance between video contents and the representative words in topics to generate better captions.
Results in Table~\ref{tab:annotate_topic_metric} and examples in Figure~\ref{fig:gt_topic_eg} presents the captioning performance with the annotated topics.
Both show that with more accurate topic information the captioning performance can be further improved.

\begin{figure} \centering 
\includegraphics[width=0.9\linewidth]{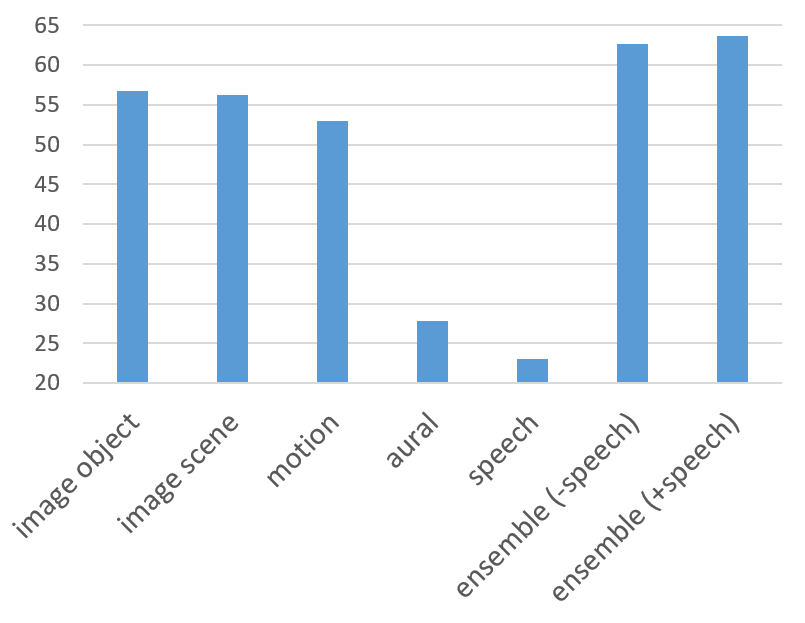} 
\caption{Accuracy of topic prediction with different modalities and ensembles. (-speech) and (+speech) means ensemble without and with speech modality.} 
\label{fig:topic_pred_acc} 
\end{figure}

\textbf{Influence of Multi-Modalities:}
As an implicit evaluation, we can evaluate the topic prediction performance with the annotated topics on testing set.
The prediction accuracies with different modalities are shown in Figure~\ref{fig:topic_pred_acc}.
The performance of the aural and speech modality are evaluated only on videos containing the corresponding modality.
Though the aural modality alone do not perform well, the ensemble of aural with image and motion modalities improves the prediction significantly with an absolute 6\% boost.
By further using the speech modality predictions, the accuracy is improved from 62.61\% to 63.65\% on testing set.

To explicitly explore the usefulness of speech modality in video captioning, we use two kinds of predicted topics which are obtained by the fusion with or without the predictions from speech modality.
Results are presented in Table~\ref{tab:speech_topic_metric}.
We can see that the captioning performance achieves large gain in all metrics although the speech modality gets only 1.04\% absolute prediction accuracy improvement as mentioned above.
It mainly results from the similar topic probability distributions using the shared topic mining model for speech modality.
So the speech modality is quite useful to generate topic proposals and thus boost the captioning performance.

\begin{table}
\centering
\caption{Captioning performance with or without using speech modality for topic prediction. The acronyms B, M, R, C denote BLEU$@$4, METEOR, ROUGE-L and CIDEr respectively.}
\label{tab:speech_topic_metric}
\begin{tabular}{c|c|c|c|c}
& B & M & R & C\\ \hline
w/o speech & 0.4266 & 0.2910 & 0.6192 & 0.4874 \\
with speech & \textbf{0.4397} & \textbf{0.2921} & \textbf{0.6234} & \textbf{0.4970}\\ 
\end{tabular}
\end{table}

\textbf{Influence of the Number of Topics:}
We also explore the performance of TGM with different numbers of topics.
As shown in Table~\ref{tab:num_topics}, the number of topics 20 achieves the best performance, which is the balance between the topic prediction performance and the topic guidance performance.
When there is fewer number of topics, the accuracy of topic prediction is higher but it provides less guidance to generate video descriptions.
When there is more number of topics, though the topic guidance becomes strong, the captioning performance suffers from the low topic prediction accuracy.
Since the more the topics are the more the topics resemble semantic concepts, it suggests that using a small number of topics is enough for the video captioning task and are superior to the large number of detected concepts. 

\begin{table}
\centering
\caption{Captioning performance with different numbers of topics. The acronyms B, M, R, C denote BLEU$@$4, METEOR, ROUGE-L and CIDEr respectively.}
\label{tab:num_topics}
\begin{tabular}{c|c|c|c|c}
\#topics & B & M & R & C\\ \hline
10 & 0.4258 & 0.2889 & 0.6196 & 0.4789\\
20 & \textbf{0.4397} & \textbf{0.2921} & \textbf{0.6234} & \textbf{0.4970}\\ 
30 & 0.4229 & 0.2901 & 0.6169 & 0.4799 \\
\end{tabular}
\end{table}

\section{Conclusions}
Descriptions of videos with diverse topics vary a lot in vocabularies and expression styles. 
In this paper, we propose a novel topic-guided model to deal with the topic diversity nature of videos. 
It can generate better topic related descriptions for videos in various topics.
Our experimental results show that the topic information is very useful to guide the caption model for more topic appropriate description generation and topics automatically mined in data-driven way are superior to the predefined topics as the topic guidance.
Multimodal features especially the speech modality features are vital to predict topics for testing videos.
Our proposed topic-guided model which functions as an ensemble of topic-aware language decoders can utilize the topic information more effectively than other caption models. 
It significantly improves the multimodal baseline performance on the current largest video caption dataset MSR-VTT, outperforming the winning performance in the 2016 MSR video to language challenge.
In the future work, we will continue to improve the topic prediction performance and jointly learn the topic representation and caption generation end-to-end.

\section{Acknowledgments}
This work is supported by National Key Research and Development Plan under Grant No. 2016YFB1001202.

\bibliographystyle{unsrt}
\balance
\bibliography{reference} 

\end{document}